\documentclass[letterpaper, 10 pt, conference]{ieeeconf}  

\IEEEoverridecommandlockouts                              

\usepackage{graphics} 
\usepackage{epsfig} 
\usepackage{times} 
\usepackage{amsmath} 
\usepackage{amssymb, amsfonts}  
\usepackage{enumerate}
\usepackage{algorithmic}
\usepackage{cite}
\usepackage{multirow}
\usepackage{subfig}
\usepackage[ruled,vlined]{algorithm2e}
\usepackage{fancyhdr,graphicx,amsmath,amssymb}
\usepackage{bbm}
\usepackage{hyperref}
\usepackage{xcolor}
\usepackage{caption}
\usepackage{dsfont}

\title{\LARGE \bf
Survivable Robotic Control through Guided Bayesian Policy Search with Deep Reinforcement Learning
}

\author{Sayyed Jaffar Ali Raza$^{\ast}$, Apan Dastider$^{\ast}$, and Mingjie Lin$^{\dag}$\\
\thanks{\newline $\ast$: Equal contribution. Correspondence: \url{jaffar@knights.ucf.edu} \newline\dag: Mingjie Lin is associate professor with the Department of Electrical and Computer Engineering, Univ. of Central Florida, Orlando, FL 32826, USA }
}%

\begin{document}
\maketitle
\thispagestyle{empty}
\pagestyle{empty}

\begin{abstract}
Many robot manipulation skills can be represented with deterministic characteristics and there exist efficient techniques for learning parameterized motor plans for those skills. However, one of the active research challenge still remains to sustain manipulation capabilities in situation of a mechanical failure. Ideally, like biological creatures, a robotic agent should be able to reconfigure its control policy by adapting to dynamic adversaries. In this paper, we propose a method that allows an agent to survive in a situation of mechanical loss, and adaptively learn manipulation with compromised degrees of freedom--- we call our method \textit{Survivable Robotic Learning} (SRL). Our key idea is to leverage Bayesian policy gradient by encoding knowledge bias in posterior estimation, which in turn alleviates future policy search explorations, in terms of sample efficiency and when compared to random exploration based policy search methods. SRL represents policy priors as Gaussian process, which allows tractable computation of approximate posterior (when true gradient is intractable), by incorporating guided bias as proxy from prior replays. We evaluate our proposed method against off-the-shelf model free learning algorithm (DDPG), testing on a hexapod robot platform which encounters incremental failure emulation, and our experiments show that our method improves largely in terms of sample requirement and quantitative success ratio in all failure modes. A demonstration video of our experiments can be viewed at: \textcolor{cyan}{\url{https://sites.google.com/view/survivalrl}}
\end{abstract}

\begin{keywords}
Survivable Robotic Control, Bayesian Learning, Guided Policy Search
\end{keywords}
\section{INTRODUCTION}
Reinforcement learning (RL) have recently demonstrated promising performance in domain of robotic learning, solving problems like robot manipulation ~\cite{robo-manip}, motion planning ~\cite{raza_srl} and robot control ~\cite{Lillicrap2016}. Moreover, RL methods for robotics are expected to shift the automation paradigm by exhibiting high tolerance or error resilience, thus capable of autonomously accomplishing tasks under complex environments. However, such overwhelming vision equally demands reduced training overhead and tractable computational requirements. One of the key challenges for RL algorithms is there requirement of extensive interaction with the environment, gathering experience to train policies that solve new tasks ~\cite{ac-teach}. This requirement, in particular grows significantly in domain of high-dimensional robotic systems where gathering experience from interactions is slow and expensive. Moreover, the gathered experience could also become less relevant if the composition of robotic system varies over time---hence requiring to regather the samples ~\cite{MC-unsound, goal_based_planning}. 
\begin{figure}[]                                       
    \centering
    
    \captionsetup{justification=justified}
    \includegraphics[width=\linewidth]{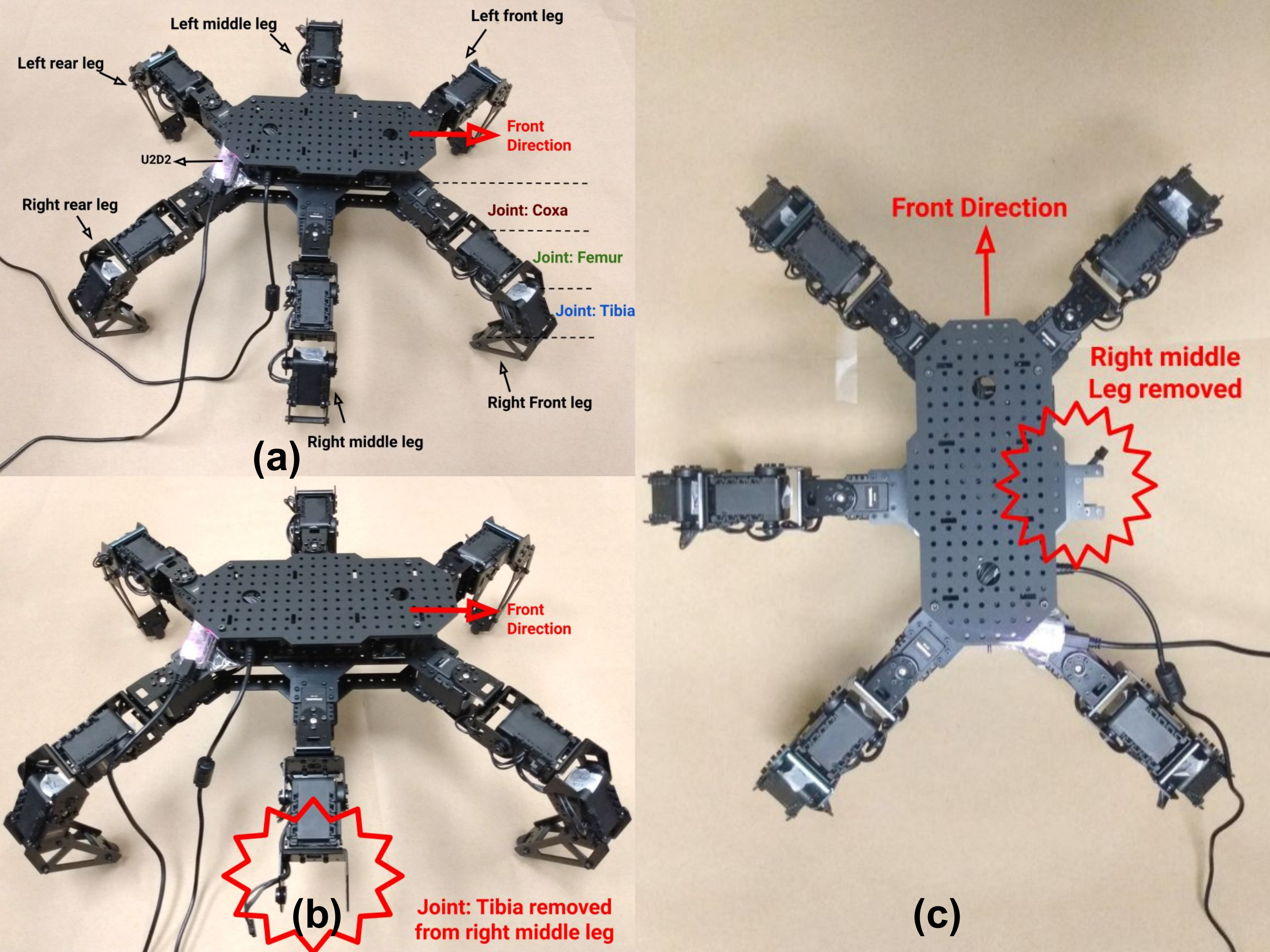}  
    \caption{(a):18-DoF Hexapod used for experiments (b):Failure shown as reduction in one DoF (c):Failure case where entire limb is amputated, reducing three DoFs as well as the mass. }%
    \label{fig1:intro-fig}                                   
\end{figure}
Intuitively, a possible approach to address this challenge could be defined by representing past gathered samples as a generative representation of the model being learnt, allowing the agent to reproduce or bootstrap parameters from past experiences to alleviate sample requirements for learning policy for future tasks. We study about how efficiently an agent can adapt its learning towards task completion, when encountered with adversaries like physical damage to joints or random loss in degrees of freedom, mimicking mechanical failure. Our proposed {\em Survivable Robotic Learning} (SRL) framework enables the robot to learn and update its future policy by inducing bias from prior behavioral samples. Given a learned policy $\pi$ for a deterministic robotic agent Fig.\ref{fig1:intro-fig}.a, our method aims at quickly adapting to sudden changes in system dynamics or robotic agent itself as shown in Fig.\ref{fig1:intro-fig}.b and Fig.\ref{fig1:intro-fig}.c respectively; for example, an agent in Fig.\ref{fig1:intro-fig}.a is deterministically trained, and is equipped with control policy $\pi$ assuming complete functionality, but unexpectedly looses certain portion of its mobility or faces with unforeseen occurrences; now would need to relearn a policy separately in case of Fig.\ref{fig1:intro-fig}.b and Fig.\ref{fig1:intro-fig}.c. Mathematically, all these scenarios can be formally abstracted as the loss in state space $\mathcal{S}$ or action space $\mathcal{A}$. As such, the goal of our SRL is to compute a new optimal policy $\pi_{new}$ considering the changes in $\mathcal{S}$ and $\mathcal{A}$, while doing generative exploitation of previously learned samples from past policies.

Our key idea is to use parametric posterior estimation (Bayesian RL) to guide the search direction of new policy while exploiting the base policy as much as possible inspired by \cite{Ghavamzadeh2016}. Technically, our proposed SRL exploits Bayesian exploration strategy to wisely guide the exploration of on-going learning policy by integrating the past learned knowledge through prior distributions. Bayesian learning not only stabilizes exploration-exploitation dilemma during gradient estimation, but also tackles the central challenges of the explosion of sample-complexity imposed by off-the-shelf Monte-Carlo based policy gradient approaches (~\cite{Lillicrap2016,MCPG,MC-unsound}). Thus, we can compute policy gradient and policy integral (posterior) with Bayesian gradient estimation\cite{Ghavamzadeh2007} and the samples of its integrand distributions(a priori), while avoiding the high parametric divergence of MC based update.

\section{Related Work}
In this section, we will briefly discuss about existing work that utilize RL methods for robotic applications, specially achieving robotic control using probabilistic methods for learning control policy through Bayesian estimation methods that use prior distributions to bias parametric updates in posterior inference.


\textbf{{\em Bayesian Reinforcement Learning}---} Bayesian methodology, which incorporates prior information into inference algorithms, has been extensively studied to augment the conventional reinforcement learning (RL)~\cite{Ghavamzadeh2015}. Such integration  provides an elegant approach to account for prior knowledge and learning uncertainty while effectively trading between exploration and exploitation during action-selection. Although the majority of recent literature on Bayesian RL is model-based~\cite{Ghavamzadeh2015} (priors are expressed over the value function or policy class), in robotics, because prior information rarely can be accurately expressed as a parameterized Markov model~\cite{Ghavamzadeh2016,Ghavamzadeh2007}, this paper will exclusively focus on Bayesian methods for model-free RL, where priors are expressed over the value function or policy class.  Specifically, the study in~\cite{Ghavamzadeh2007} proposed  a new actor-critic learning model consisting of a Bayesian class of non-parametric critics using Gaussian process temporal difference learning. Recently in~\cite{Ghavamzadeh2016}, Bayesian framework has been shown that modeling the policy gradient as a Gaussian process can significantly improve the efficiency of sample usage and acquire uncertainty measurements almost free.

\textbf{{\em Damage recovery and sudden perturbation planning in robotics}---} Previously, to mitigate the adverse effect due to sudden malfunction, most robotics systems were pre-loaded multiple contingency planes constructed beforehand. As such, as soon as robotic damage occurs, the robot would first conduct self-diagnosis, quickly assess its damage, and subsequently select the best recovery plan\cite{rtfault_diganosis}. Unfortunately, such self-diagnosing robots are overly challenging to design, because foreseeing all possible failure modes is extremely difficult and providing all backup plans can be prohibitively expensive. For example, in~\cite{Bongard2006}, a 4-leg robot has been demonstrated to recover from a disabled leg autonomously without pre-stored contingent policies by continuous self-modeling. The main algorithm used in~\cite{Bongard2006} is based on the basic stochastic optimization that attempts to explain the observed sensory-actuation causal relationship, therefore largely heuristic-based. More recently, in~\cite{Cully2015}, an evolutionary algorithm has been developed that allows a six-legged robot to adapt to unforeseen changes. The six-legged mobile robot investigated used an intelligent trial-and-error algorithm to tap into the experiences previously accumulated and quickly find optimal compensating behaviours. 

\textit{{\bf Key SRL differences---}} Compared with the prior studies, our proposed SRL algorithm has at least two fundamental distinctions. First, both~\cite{Bongard2006} and~\cite{Cully2015} applied control system-based methodology to tackle damage recovery in robotics and these methods have been quite successful in various domains, but here we wanted to exploit the most recent development in deep RL for robotics. This is critical because deep RL is proved not only to provide a mathematically rigorous machine learning framework, but also is more accurate to account for the continuous state and action space for robotic control. Second, although the study~\cite{Cully2015} did utilize a Bayesian-inspired prediction scheme, our proposed methodology tightly incoorperate the Bayesian learning algorithm with the most recent off-model and off-policy deep reinforcement learning, thus much more capable and more sophisticated. In many aspects, the Intelligent Trial and Error algorithm (IT\&E) proposed in~\cite{Cully2015} represent a rudimentary form of basic tabular-based Q-learning~\cite{Barto2019},which is typically used for finite state and action spaces, while most of the robotic platforms deal with continuously varying spaces. 

\section{Preliminaries}
We consider the standard RL setting represented as an MDP (Markov decision process). We represent MDP node as a tuple $(s, a, p_{0}, r, p, \gamma)$. An agent interacts with an stochastic environment consisting of a set of states $s\in\mathcal{S}$, a set of actions $a\in\mathcal{A}$, a distribution of initial states $p(s_0)$, a reward function $r : \mathcal{S} \times \mathcal{A} \rightarrow \mathbb{R}$, transition probabilities  $p(s_{t+1} | s_t, a_t) : s \times a \rightarrow s^\prime$, and a discount factor $\gamma \in [0, 1]$. The agent interacts with the environment to learn a policy $\pi (a_t | s_t)$ which is actually a mapping function representing probability distribution function (PDF) of reward, taking state and action as input random variables. An agent chooses an action according to $\pi (a_t | s_t): \mathcal{S} \rightarrow \mathcal{P} (\mathcal{A})$ such that long-term expected sum of rewards, $J = \mathbf{E}_{\pi}[\sum_{t=0}^{\infty} \gamma^t r(s_t, a_t)]$, can be maximized.

The quality of each action $a_t$ sampled by current policy $\pi$ in a state $s_t$ can be measured by a function $Q(s_t, a_t) = \mathbb{E}_{\pi} [J | s_t, a_t]$. The sequence of state-action pairs in an episode creates a trajectory $\xi = (s_0,a_0,s_1,a_1,...,s_{T-1},a_{T-1},s_g)$ in form of a Markov chain. The PDF of such Markov chain followed by policy $\pi$ is represented as $P(\xi_\pi)=p(s_{0})\prod_{t=0}^{T-1}\pi(a_{t}|s_{t})p(s_{t+1}|s_{t},a_{t})$, which is generated by Monte-Carlo sampling. Also, the expected return for a given $\xi$ can be expressed as, $\eta(\pi)=\mathbf{E}[J(\xi_{\pi}))] = \int J(\xi_{\pi})P(\xi_{\pi})d\pi$.

\subsection{Bottlenecks in Monte-Carlo Simulation}
The policy search is performed by estimating the gradient over expected return
from a class of parameterized stochastic policies $\{\pi(\cdot| s, \theta),
s \in \mathcal{S}, \theta \in \Theta\}$ w.r.t policy parameters $\theta$ from
observed system trajectories and then improve the policy by adjusting the
parameters in the direction of gradient. For, any given set of trajectories $\xi_{1},\xi_{2},\dots,\xi_{M}$ we can
state Monte-Carlo (MC) estimator as
$  \widehat{\nabla \eta}_{MC}(\pi_{\theta}) =\frac{1}{M} \sum_{i=1}^{M} 
J(\xi_{i}) \nabla \log P(\xi_{i}| \pi_{\theta}) 
=\frac{1}{M} \sum_{i=1}^{M} J(\xi_{i}) \sum_{t=0}^{T_{i-1}} \nabla \log \pi(a_{t, i} \mid s_{t, i}, {\theta})
$, where gradients are defined as likelihood ratio. So, in MC based estimation, policy gradients are defined over the expected values of trajectories and the samples are drawn according to their probability values. This policy gradient can be an unbiased estimate of true gradient $\eta(\cdot)$ only when number of drawn samples tend to infinity i.e $M\rightarrow\infty$ and thus, variance between $\widehat{\nabla \eta}_{MC}(\pi_{\theta})$ and true gradient $\nabla\eta(\pi_{\theta})$ diminishes to zero. This condition makes MC based policy modeling infeasible for robotic platforms, because gathering experiences in robot learning is slow and expensive. Moreover, MC estimation is fundamentally unsound as explained in \cite{MC-unsound}, because the estimator only depends on the values of sampling
distribution which are arbitrary choices, and are dependent on stationary
reward distributions. This dependence violates the likelihood principle because
the estimated gradient becomes irrelevant if the reward distribution is likely
to evolve with time \cite{Ghavamzadeh2007}. Therefore, for problems with temporally varying distribution,
classical policy gradient alone would yield non-optimal performance.  

\subsection{Gradient Computation by Bayesian Quadrature}

RL for robotic control typically deals with continuous state and
action spaces with extremely high dimensionality, and environments and the associated robotic agent are largely susceptible to uncertain changes due to inherent non-stationarity in work-space and unwanted malfunction in the agent's structure itself. In Bayesian RL, the distribution of unknown function $f(\cdot)$ which can be a representative for robust control policy $\pi(\cdot|s_{t})$ to tackle uncertainty mentioned earlier, is modelled as a Gaussian Process (GP) by defining a Gaussian distribution as prior distribution over functions. The inferred posterior distribution from this prior will also be a Gaussian normal.  Therefore, the gradient of expected
return in terms of Bayesian quadrature is given by $\eta_{B}({\pi_\theta})=\int
J(\xi_\pi) P(\xi_\pi, {\theta}) d \xi$. We consider
$\eta_{B}({\pi_\theta})$ as a random variable due to high variance (Bayesian
uncertainty) in $J(\xi_\pi)$. As such, the expected mean of a posterior distribution of
gradient is computed as
\begin{equation}
    \label{eq4:bayesian-pg}
    \begin{aligned}
    \nabla \mathbf{E}[\eta_{B}({\theta}) \mid \mathcal{D}_{M}] &=\mathbf{E}[\nabla \eta_{B}({\theta}) \mid \mathcal{D}_{M}]  \\
    &=\mathbf{E}[\int J(\xi_\pi) \frac{\nabla  P(\xi_\pi ,{\theta})}{P(\xi_\pi , {\theta})} P(\xi_\pi , {\theta}) d \xi \mid \mathcal{D}_{M}],
    \end{aligned}
\end{equation}
where the set of samples $\mathcal{D}_M$ 
${={\{(s_{i},\hat{s}_{i})\}_{i=1}^{M}}}$ are provided and 
$\hat{s}$ is the noisy
observation of $f(s):s\times a \to s^\prime$. The integrand in Eq.
\eqref{eq4:bayesian-pg} can be decomposed into a GP prior function
$f(\xi_\pi,\theta)$ and its probability distribution $p(\xi_\pi,\theta)$. When
computing posterior, the quadrature assumes $p$ to be known. 

Despite of addressing the issue of high variance,
calculating the gradient in Eq. \eqref{eq4:bayesian-pg} 
implicitly assumes that (i) the behavior of prior moments can be aggregated
into a policy distribution which is a normal Gaussian process and (ii) the
parameter $\theta$ of prior can be embedded for determining direction of
posterior gradient. Therefore, we augmented the policy learning with Bayesian optimization for exploiting the past experiences and prior policy gradient updates to guide new policy search in order to handle sudden perturbations in state-space or action-space of the robotic agent.

\section{Proposed Approach}
The major objective of our proposed SRL is enabling a robotic agent to inherit, partially or entirely, a previously learned motion policy in order to quickly evolve or adapt to a new policy that  optimally tackles unexpected sudden changes in state dynamics, such as joint damages during run time. The key intuition behind our SRL, which has been biologically confirmed as well~\cite{Bongard2006}, is that, even if a robot is damaged, its prior motion policy $\pi_B$ can still be partially utilized to characterize behavior for both initializing the parameter vector for $\pi_{new}$ and determining the direction for making gradient updates $\nabla \pi_{new}$. Abstractly, to formalize our SRL formulation, we defined two separate domain of operation--{\em healthy domain} and {\em unhealthy domain}. At the beginning, our SRL agent was assumed to be equipped with a well-learned policy $\pi_B$ and a healthy model $\mathcal{M}_B$, where the subscript $B$ represents \textit{behavior}--- as we will use behavior policy to learn target (new) policy. In our case, the robot was initially intact with all available degree of freedoms (DoFs) and zero malfunction. The knowledge learned in this {\em healthy domain} can be encoded as a heuristic ensemble guide that directs the exploration and parameter updates for future {\em unhealthy domain}. Hence, learning in a new domain need not to be initiated from absolute scratch, rather our SRL agent strives to learn unknown tasks through performing a guided exploration to explore states unknown to {\em healthy domain}.  
\begin{figure}[htbp]                                       
    \centering                                             
    \includegraphics[width=\linewidth]{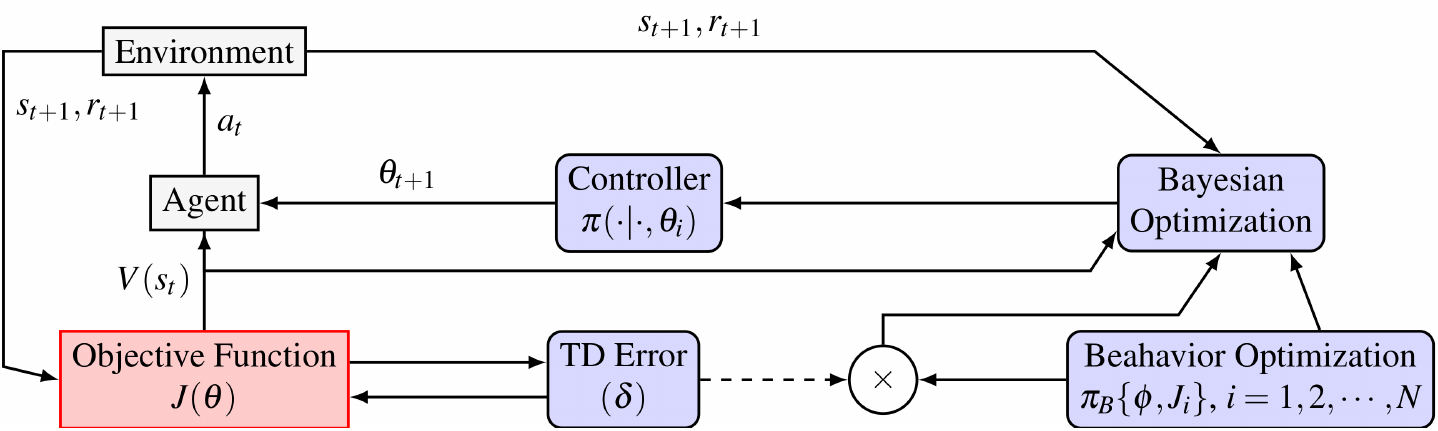}  
    \caption{Overall learning framework of SRL.}%
    \label{fig:overall}                                   
\end{figure} 
When our SRL agent suffered from unknown functionality loss in terms of its mobility remarked as unhealthy model $\mathcal{M^+}$, naturally the previous healthy policy $\pi_B$ would under-perform in $\mathcal{M}^+$ due to sudden changes in state space and action-space. However, instead of totally disregarding the learning trends in previous $\mathcal{M}_B$, our SRL agent can exploit its prior locomotion policy by estimating state value similarity between known $\mathcal{M}_B$ and unknown $\mathcal{M}^+$ to quickly plan new locomotion plans for maneuvering in new {\em unhealthy domain}. Fig. \ref{fig:overall} depicts the overall learning strategy for our SRL agent. As such, we view the SRL problem as an incremental estimation, where a new trajectory is suggested by the $\pi_{new}$ considering the old behaviors of the agent as a set of advise to plan locomotion in the $\mathcal{M}^+$ domain. In short, the objective of the SRL state estimation is to find a continuous-valued posterior trajectory for $\mathcal{M}^+$ given an ensemble prior distribution optimized by policy in $\mathcal{M}_B$.       
\begin{figure}[htbp]                                       
    \centering                                             
    \includegraphics[width=\linewidth]{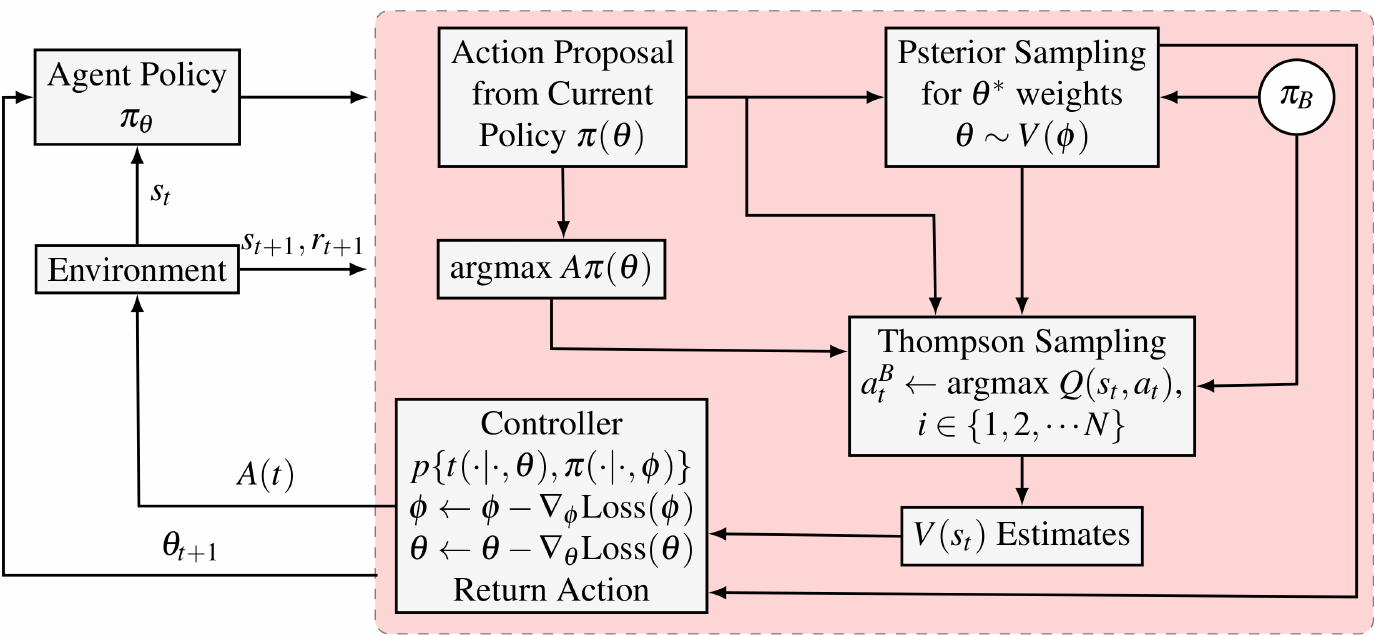}  
    \caption{Algorithmic details for parametric updates in SRL}%
    \label{fig:algo}                                   
\end{figure} 

\subsection{Behavior Policy: Healthy Ensembles}
To formalize healthy ensemble attributes, we consider infinite-horizon MDPs with stochastic transition dynamics $p(s_{t+1}|s_t, a_t)$, a set of terminal states $\mathcal{G}_{B} \subset \mathcal{S},$ and a sparse reward function $r_{\mathcal{G}_{B}}(s_{B}) =\mathds{1}(s_{B} \in \mathcal{G}_{B})$. For any stochastic behavior policy $\pi_{B}$, the state value function is $$V_{\mathcal{G}_{B}}^{\pi_{B}}\left({s_{(0,B)}}\right)=\mathbf{E}\big[\sum_{t=1}^{\infty}\gamma^{t} r_{\mathcal{G}_{B}}\left(s_{t,B}\right)\big]$$ $B$ superscript specifies that these information are collected from prior domain $\mathcal{M}_{B}$. By following behavior policy $\pi_{B}$, the agent creates a trajectory of state-action pairs as, $\xi_{\pi_B} = (s_0, a_0, s_1, a_1, ...)$ where $a_t \sim \pi_{B}(\cdot|s_t)$ and $s_{t+1}\sim p(\cdot|s_t,a_t)$. Each trajectory either ends by reaching a goal state $s_{g}$ or exhausts all allowed steps and fails to reach a goal state. All trajectories can be clustered into two sets $\mathcal{X}_{g}$ and $\mathcal{X}_{g}^{C}$ such as, all successful trajectories $\xi_{i_{\pi_{B}}} \in \mathcal{X}_g$ and all failure states $\xi_{j_{\pi_{B}}} \in \mathcal{X}_g^C$ (where $C$ represents \textit{cancelled} trajectories). For any goal state $s_{{g}_{B}} \in\mathcal{G}_{B}$ , $s_{{g}_{B}}$ is absorbing with reward $1$, then the value for $s_{{g}_{B}}$, is $V^{\pi_{B}}\left(s_{{g}_{B}}\right)=\frac{1}{1-\gamma}$. 
Now, any trajectory $\xi_{i_{\pi_{B}}} \in \mathcal{X}_g$ that ends being successful within allowed time-steps $t_g$, generates a return of $\frac{\gamma^{t_{g}-1}}{1-\gamma}$ and all trajectories belonging to $\mathcal{X}_g^C$ returns always $0$. Therefore, the expected value function expression can be written as expected discount sum of rewards for a trajectory $\xi_{\pi_{B}}$ as,
\begin{align}
    \label{eq1:exp_val}
    V_{\mathcal{G}_{B}}^{\pi_{B}}\left(s_{(0,B)}\right)&=\mathbf{E}_{\xi}[R(\xi_{\pi_{B}})] \nonumber \\
    &=P(\xi_{\pi_{B}}\in \mathcal{X}_g)\mathbf{E}_{t_g}\big[\frac{\gamma^{t_g-1}}{1-\gamma}\big] \nonumber
 \end{align}
The above expression of value function contains two important attributes: (1) reaching a goal state from state $s_{(0,B)}$ with higher certainty and (2) reaching the goal state quickly \cite{ac-teach, rl_book}.

So when we build the prior healthy ensembles taking into consideration the actions taken by policy $\pi_{B}$ and the effects of those actions while computing the value functions for any state $s$, we are taking two-fold advantages from prior domains--how much probable this trajectory is for reaching a goal state and how fast it converges to optimal solution, i.e completion of task with highest return possible.  Thus, the value functions can be utilized to extend the ensemble attributes, since value functions implicitly carry information about past successful trajectories the agent created. 

\subsection{Behavior Policy: Partially Guidance}
Based on the {\em goal reaching probability and time} based healthy ensemble setting discussed above, the prior behavior policy, in some known regions, would transit from one state to another state as an optimal policy would reach quickly the goal state from the new state. Thus, in that known region,
the behavior policy would offer partially useful advice for the target policy $\pi^{+}$ to tackle unhealthy domain $\mathcal{M}^{+}$.
Let $\pi^{*+}$ denote the optimal policy and $V_{\mathcal{G}}^{*+}(s)$ denote the optimal
state value function, with respect to a fixed set of goals $\mathcal{G^+}$.
Then the behavior policy can only guide partially, if in some non-empty strict
subset $s_{g} \subset \mathcal{S}, \forall s_{g} \in \mathcal{S},$ we have that $V_{\mathcal{G}}^{*+}(\xi_{B}(s_{g},a))>V_{\mathcal{G}}^{*+}(s_{g})$, where $\xi_{B}$ denotes the transition generated with behavior policy $\pi_{B}$. This is also visually depicted in Fig.~\ref{fig:algo}, showing that the partial guidance from $\pi_B$ acts as behavior ensemble and helps the agent to model the future policy fast by utilizing partial information of behavior samples and computing $\nabla \pi_{new}$ gradient in direction of prior ensembles.

\subsection{Thompson Sampling}
Exploitation of healthy ensemble needs to be balanced between those states whose utility is non-stationary or unknown to behavior policy $\pi_{B}$. We adapt Thompson Sampling~\cite{thompson_sampling}, which is widely accepted Bayesian method for maintaining balance between exploration and exploitation, by modeling the uncertainty as a posterior distribution for each policy. Inspired by~\cite{ac-teach}, we use Thompson sampling to model posterior distribution over expected state-action values within the critic network. The reason behind maintaining posterior distribution, instead of $q \, \forall\, Q_{\phi}$ estimates, the distribution is maintained over network weights (parameters) and consequently over Bayesian confidence. The weights are forward passed to the controller (as shown in Fig.~\ref{fig:algo}), which then decides what action to take by evaluating action proposals and corresponding values.

\subsection{Controller}
It is potentially possible that the behavior policy $\pi_{B}$ can distract our SRL agent from optimal learning, especially in those state-action choices which are unseen by $\pi_{B}$. Additionally, there is also a chance that our SRL agent formulates a virtual local minima, and switches excessively between $V(\phi) \sim \pi_{B}$ and $V(\theta) \sim \pi_{\theta}^{*+}$. To mitigate these risks, we setup a controller (see Fig.~\ref{fig:algo}), that compare the policy parameters for both $\theta_{i}$ and $\phi_{i}$, selected at timestep $t_{i}$, by performing Thompson sampling process between ($\theta_{i}$, $\phi_{i}$) and ($\theta_{i-i}$, $\phi_{i-i}$). The controller also utilizes posterior sampling to estimate the probabilistic state-action values of actions suggested by prior healthy ensemble as $\arg \max_{i\in 1,...,N} Q(s,a_{i})$, if this value is larger than actor's action proposal value, the behavioral policy acts as the new policy at timestep $i$, otherwise the agent acts with the previous policy. We present more algorithm details in Algorithm~\ref{algo}

\begin{algorithm}
\scriptsize
   \caption{SRL Bayesian Inference Algorithm}
\label{algo}
\SetAlgoLined
  \SetKwData{Left}{left}
  \SetKwData{Up}{up}
  \SetKwFunction{FindCompress}{FindCompress}
  \SetKwInOut{Input}{input}
  \SetKwInOut{Output}{output}
  \textbf{Require:}
  \\ \quad $\pi_B$ : Behaviour Policy 
  \\ \quad $S_t$  : Current Observation
  \\ \quad $\mathcal{A}$ : Action Proposal
  \BlankLine
  \While {$epoch < maxEpoch$} {
  \For{$episode=\{1,2,...,n_{episodes} \}$} {
  \For{$steps=\{1,2,...,n_{steps}\}$}{ 
  $\mathcal{A} \leftarrow [\pi_\theta (S_t)$ , $\pi^B_\phi (S_t)]$  {\scriptsize\textit{\#Observe Action Proposal}}
  \\$\theta_{t+1}\backsim q(\phi)$  {\textit{\scriptsize\# Posterior dist. sampling}}
  \\{\small$\operatorname{TS}(a_{t}) \leftarrow \arg\max_{i\in1,. . N}{Q_\phi(s,a)}$ {\scriptsize\textit{\# Thompson}}}
  \\{\scriptsize\textit{\# Get probability of higher expected value b/w two timsteps}}
  \\$p_a\leftarrow \textbf{P}[Q_\phi (S_t,\small\operatorname{TS}(a_{t}))>Q_\phi (S_t,\small\operatorname{TS}(a_{t-1}))]$  
  \\\uIf{$P_a$}{
  \CommentSty{CNTRL} : accept proposed action
  \\choose $a_t\leftarrow \mathcal{A}$
  }
  \uElse {
  $V(s_t) \leftarrow$ value update $E[G_t|s_t]$
  \\choose $\arg\max_{\pi_\theta}(V(s,a))$
  }
  \CommentSty{CNTRL} Update:
  \\ \quad Loss: $\phi \leftarrow \phi - \nabla_\phi L_{critic}(\phi)$
  \\ \quad Loss: $\theta \leftarrow \theta - \nabla_\theta L_{actor}(\theta)$
  \\ \quad $V(S_t) \leftarrow$ value update
  \\ \quad \CommentSty{return($a_{t}$)} to agent
  }
  }
  }
 
\end{algorithm}

\section{System Overview and Experiments}
In this section we describe our simulation and physical frameworks that were utilized to conduct comparative testing of SRL and DDPG~\cite{Lillicrap2016}.

\subsection{Hardware Platform} 
\label{Hardware}

As shown in Fig.~\ref{fig1:intro-fig}, 
we used a Phantomx{\textregistered} AX Metal Hexapod Mark III crawler robot
to conduct our experiments. 
It consists of six legs and each leg
comprises of three individually controlled
joints named as ``coxa", ``femur", and ``tibia".
Altogether this crawler robot comprises 18
degrees of freedom (DOF), which make it
versatile and robust for conducting our experiments.
To read and write 18 joint angles of all 6 legs of our hexapod, 
we opted to use an U2D2 USB
communication converter instead of a more conventional Arduino compatible Arbotix controller 
because 
the U2D2 allows faster data communication with the dynamixel motors attached at each joint. 
We also integrated a PID based position controller for each joint in order to follow 
closed-loop feedback commands and shift within pre-determined operating ranges.

\begin{figure*}[h]                                       
    \centering                                             
    \includegraphics[width=\linewidth]{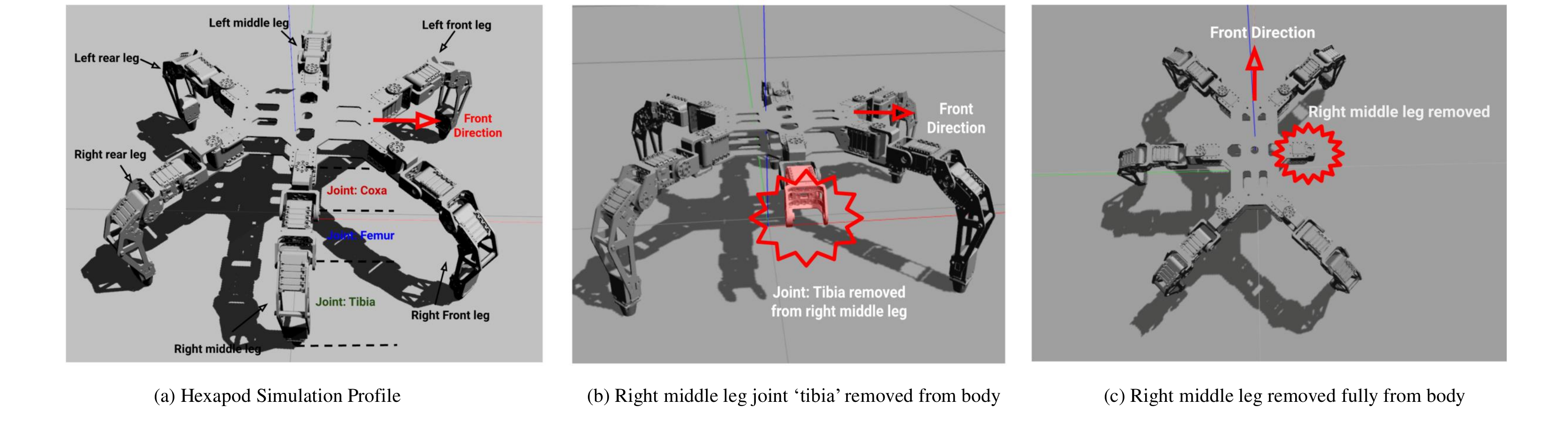}  
    \caption{Simulation model of hexapod robot and level of damages by reducing numbers of DOF.}%
    \label{fig:sim}                                   
\end{figure*}   

\subsection{Experiments overview}
\label{subsec:exp_ovrvw}
Video demo of our experiments can be viewed at: \textcolor{cyan}{\url{https://sites.google.com/view/survivalrl}}. Two pointers
were placed in the workspace of our experiments and were marked as ``start" and
``finish". 
At first, to define the baseline of our crawler's movement, our hexapod
was programmed to crawl from the start pointer to the finish pointer 
while following the tripod
gait movement pattern of a six-legged insect. 
 Afterwards, as depited in Fig.~\ref{fig1:intro-fig}, the ``tibia" joint
of right middle leg was removed manually to evaluate how the robot adapts to
its new structure and how well our proposed SRL learning algorithm performs in completing its defined 
trajectory. 
At last,
the full right-middle leg of our hexapod was removed from the body. 
As a result, our algorithm tried to
discover optimal policy to complete the full trajectory in a more complicated
circumstance of losing 3 DOF. The same length trajectory was designed in the simulation profile
for the hexapod to follow. A similar tripod gait pattern was implemented in the
simulation profile to move the robot. Eventually, one tibia joint and one full
leg was disconnected gradually from the simulation profile to evaluate the overall
performance and reward-return of our designed controller. To testify the
robustness and generalization of our algorithm and implementation, later,  different legs
were kept as a disconnected leg from the controller.

Along with random physical damage, we tested and validated SRL on three different categories of test-cases, (i) \textbf{Task\_X}, (ii) \textbf{Task\_XY} and (iii)\textbf{Task\_P2P}. 
\begin{itemize}
    \item \textbf{Task\_X}. In this easy mode, the agent's goal is to cover maximum distance along only X-axis, while its tibia joint is inoperable, i.e the robotic agent loses 1 degree of freedom as depicted in Fig. \ref{fig:sim}(b).
    \item \textbf{Task\_Y}. This is the medium difficulty mode where the agent targets to exert smooth locomotion  in 2-D environments and travels maximum distance while the one leg becomes totally damaged i.e it loses 3 DOF randomly as shown in Fig.\ref{fig:sim}(c).
    \item \textbf{Task\_P2P} In this hard difficulty mode, the objective is to arrive at specific cordinates on a planar surface, overcoming the maneuvering hindrance due to missing one full leg randomly. 
\end{itemize}
\subsection{Simulation Environment} 
\label{Simulation}
The algorithm was trained over generative sampling of state-action sequences from simulation environment.
We utilized an integrated robot simulator that combines the 3D simulator 
Gazebo and the ROS robotic interface for all of our high-fidelity simulations.
We adopted an open-source gazebo
simulation model for Phanomx AX Metal Hexapod Mark III crawler robot (as shown in Fig.~\ref{fig:sim}) 
that accurately models both the mechanical structure and the system dynamics of our target robot.
 Finally, the inter-process
communication protocol (IPC) was based on ROS ecosystem to control parallel
operation and ensure low latency feedback among simulation platform, physical
robot and our SRL controller. The Gazebo model has been spawned as an independent ROS node while its communication was synced
at the minimum rate of 20Hz to maintain handshaking message passing
protocol. 
\subsection{Algorithmic Implementation}
\label{subsec:algo_imp}
We presented SRL technique that leverages past experience as probabilistic bias towards on-going learning. The algorithmic implementation of our proposed method was significantly inspired by DDPG learning method\cite{Lillicrap2016} which is considered as one of well-established policy learning techniques for continuous robotic control. Moreover, our approach combines a probabilistic bias with conventional DDPG and computes posterior out of behavior samples---hence representing policies as Gaussian process instead of parametric mapping function. This bias enables the agent to exploit its existing policy as prior behavior ensemble and quickly learns a target policy to adapt to unforeseen domain setting. Technically, we introduced Bayesian posterior conditioned over behavior policy, that can incrementally estimate relevance between two domains, resulting in lesser exploration requirement and faster gradient updates over fewer samples.

We tested our approach on a hexapod robot with 18-DOF as shown in Fig.\ref{fig1:intro-fig}. The robot learns its behavior policy $\pi_B$ in a healthy or undamaged domain and this well-learned policy serves as the {\em healthy ensemble} to guide learning in {\em unhealthy} domain. $\pi_B$ is generic tripod gait locomotion policy which enables the agent to travel very smoothly according to three different task trajectories mentioned in section \ref{subsec:exp_ovrvw}. In the next step, we explicitly introduce adversary by amputating random joints from the robot body to emulate random {\em unhealthy} domain settings. Such random perturbations in state-space or action-space is totally unknown to $\pi_B$. In brief, the robot gets damaged and loses its maneuvering capabilities--- turning $\pi_B$ into a sub-optimal policy modeling in damaged domain. Both actor and critic network comprises of 4x1200 dense hidden layers, 1x600 hidden layer and a Bayesian dropout layer\cite{Bayesian_dropout}. We list additional training parameters in the Table \ref{tab:training_param} below:

\begin{table}[htbp]
\caption{Training Parameters}
\label{tab:training_param}
\centering
\begin{tabular}{llll}
\hline
\multicolumn{4}{c}{Training Parameters}                                                                         \\ \hline
\multicolumn{1}{l|}{Max Episodes}   & \multicolumn{1}{l|}{7,000} & \multicolumn{1}{l|}{Variance}    & $(0.9999)^t$ \\
\multicolumn{1}{l|}{Steps/Episode}  & \multicolumn{1}{l|}{2500}  & \multicolumn{1}{l|}{Learning Rate}       & $3e^{-2}$      \\
\multicolumn{1}{l|}{Replay Buffer}  & \multicolumn{1}{l|}{1600}  & \multicolumn{1}{l|}{Optimizer 1} & ADAM      \\
\multicolumn{1}{l|}{MiniBatch Size} & \multicolumn{1}{l|}{300}   & \multicolumn{1}{l|}{Optimizer 2} & RMSProp   \\ \hline
\end{tabular}
\end{table}

\begin{figure*}[htbp]
\centering
  \subfloat[]{
    \begin{minipage}[c][.5\width]{
       0.33\textwidth}
       \centering
       \includegraphics[width=1.\textwidth]{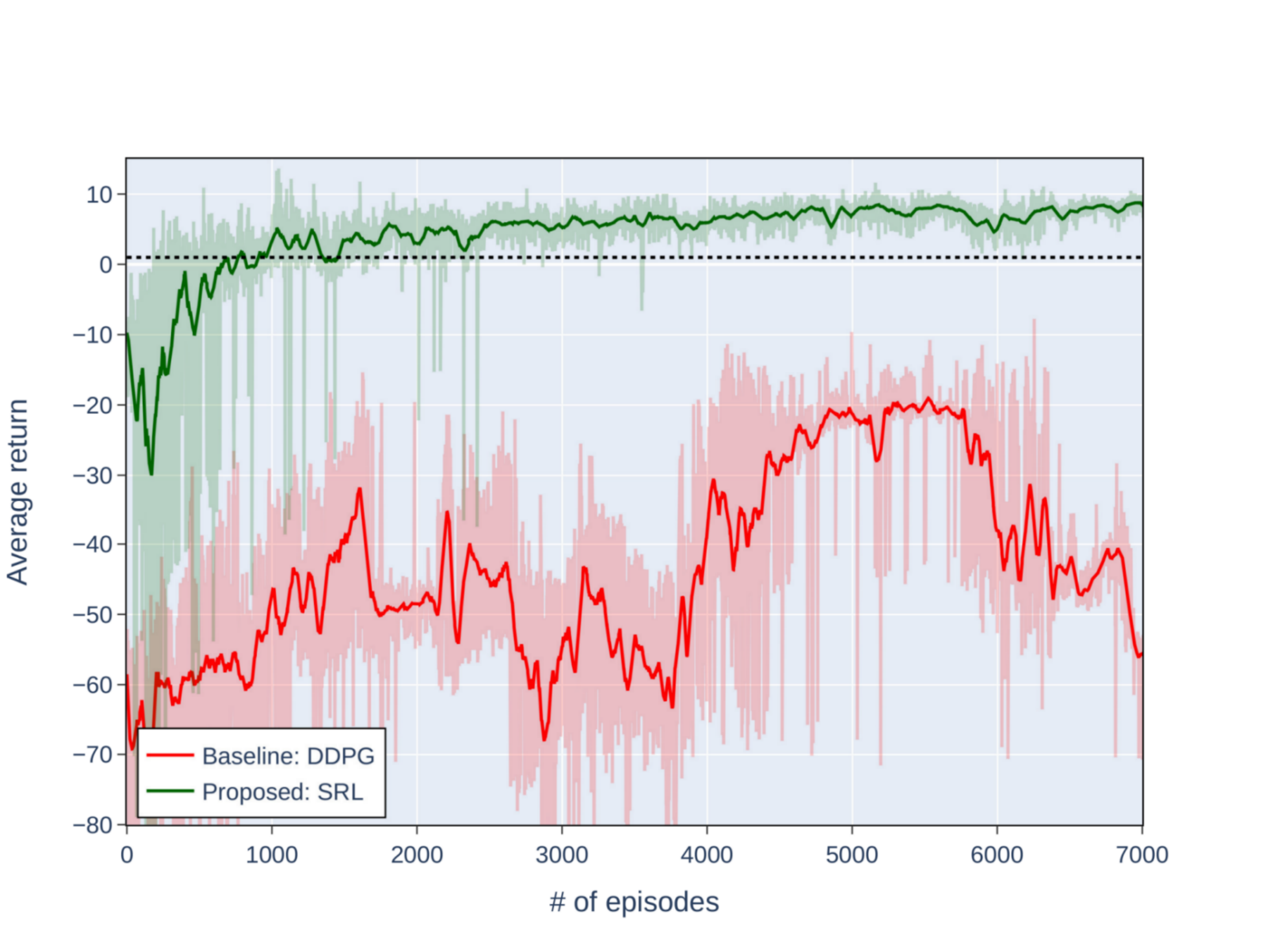}
    \end{minipage}}
  \subfloat[]{
    \begin{minipage}[c][.5\width]{
       0.33\textwidth}
       \centering
       \includegraphics[width=1.\textwidth]{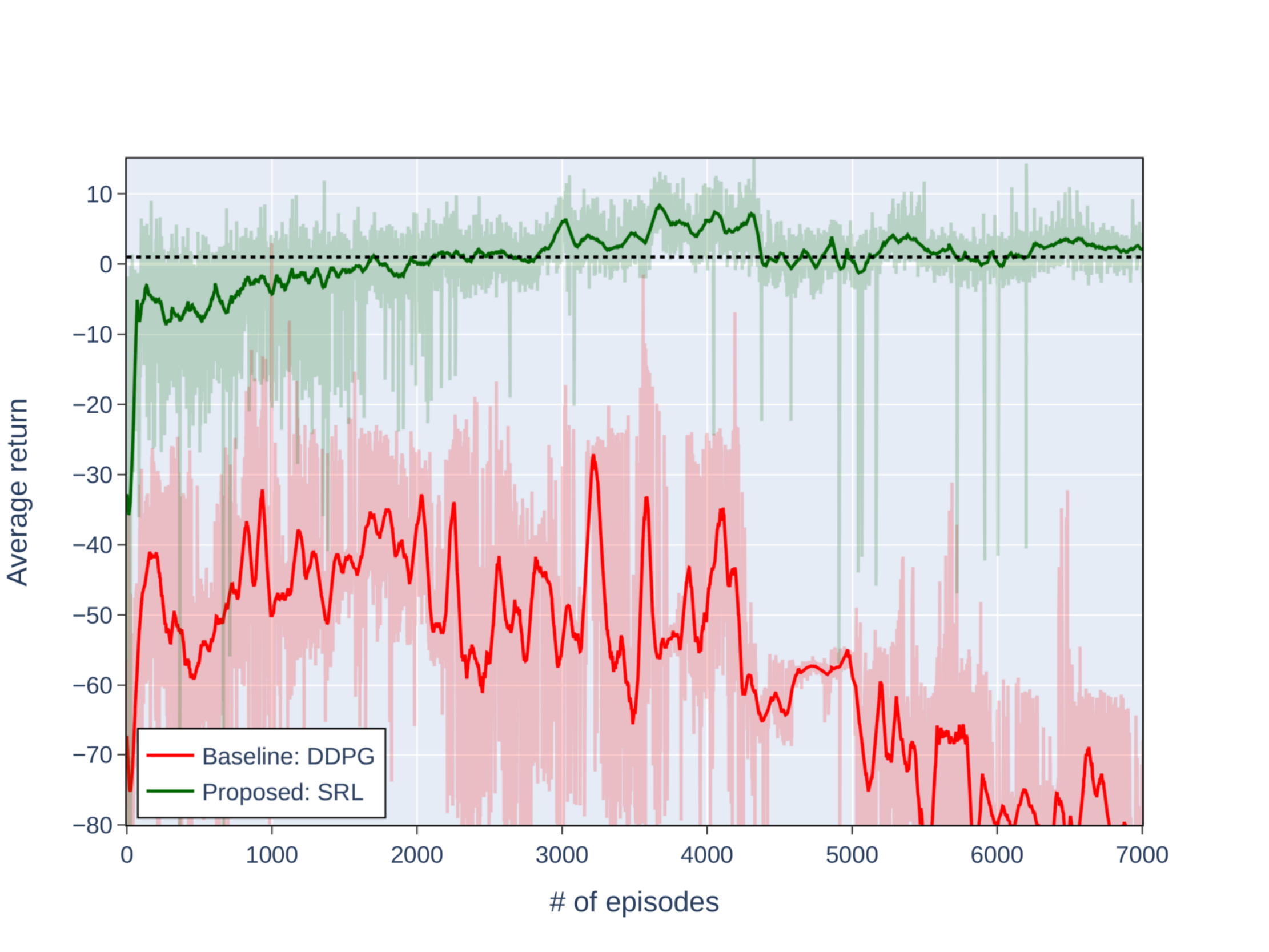}
    \end{minipage}}
  \subfloat[]{
    \begin{minipage}[c][.5\width]{
       0.33\textwidth}
       \centering
       \includegraphics[width=1.\textwidth]{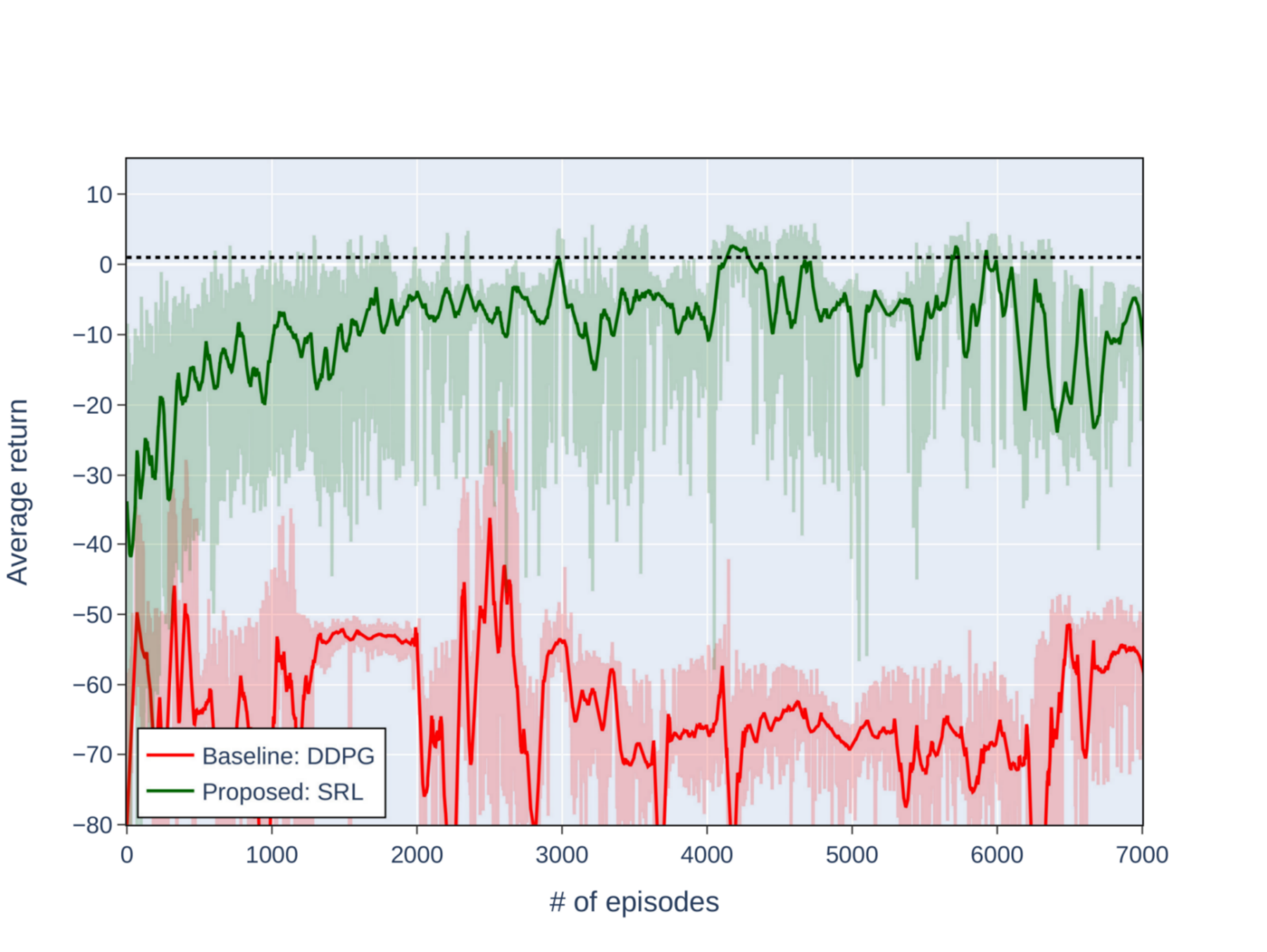}
    \end{minipage}}
	\caption{
Comparison between SRL and baseline for (a) Difficulty mode: Easy, Task: \textbf{Task\_X}; (b) Difficulty mode: Medium, Task: \textbf{Task\_Y}; Difficulty mode: Hard, Task: \textbf{Task\_P2P}.
10 random seeds are used for each test. 
}%
	\label{fig:result}
\end{figure*}

\subsection{Result Analysis}
\label{subsec:result}
Our experiments are developed to empirically investigate following questions:
\begin{enumerate}
\item How SRL can utilize the behavior policy and healthy ensemble knowledge to infer posterior distribution as target policy?
\item How efficient SRL is when new domain differs significantly from prior ensemble settings? 
\end{enumerate}

As shown in Fig. \ref{fig:result}, our proposed SRL methodology shows improved performance w.r.t sample efficiency (lesser episodic requirements), depicted as episodic returns in all three task scenarios compared to the baseline method. It can also be observed by looking at the Fig. \ref{fig:result} that standalone DDPG not only struggles to achieve optimality, but also it dips as exploration, in turn learning rate, narrows down with increasing number of episodes, trapping the baseline trend into local minima. Moreover, our Bayesian augmented SRL promisingly stabilizes the learning and exhibits less variance than the baseline. The dotted line in the plots define oracle reward threshold. This threshold can be marked as a finishing point, if reached, then an episode ends with success. We can observe from the Fig. \ref{fig:result} that the SRL method is more effective than the baseline for handling sudden disturbances occurred to state-space of a robotic agent learning optimal policy to preserve functionality.

\section{Conclusion}

Robotic systems suffer from so called ``policy fragility", meaning
that a learned robotic control policy typically can not effectively adapt to sudden changes in working environment or robotic agent itself. In sharp contrast, most living animals can quickly recover and find  a compensatory behaviour when they are injured. Partially inspired by this observation,  we propose the Survivable Reinforcement Learning (SRL) framework in order to construct optimized motion policy for a robotic agent, through integrating Bayesian priors as guided bias for future policy learning. We demonstrate promising performance of our proposed approach as an application for robotic agents working in a dynamically constrained environment and encountering random failures. In future research, we would like to apply the proposed SRL algorithm for multi-robot agents with high DoFs operating in same workspace with dynamic constraints and investigate their autonomous adaptive capability with uncertain physical damages to any agent operating in the environment. 
\bibliographystyle{IEEEtran}
\bibliography{bib.bib}

\begin{thebibliography}{10}
\providecommand{\url}[1]{#1}
\csname url@samestyle\endcsname
\providecommand{\newblock}{\relax}
\providecommand{\bibinfo}[2]{#2}
\providecommand{\BIBentrySTDinterwordspacing}{\spaceskip=0pt\relax}
\providecommand{\BIBentryALTinterwordstretchfactor}{4}
\providecommand{\BIBentryALTinterwordspacing}{\spaceskip=\fontdimen2\font plus
\BIBentryALTinterwordstretchfactor\fontdimen3\font minus
  \fontdimen4\font\relax}
\providecommand{\BIBforeignlanguage}[2]{{%
\expandafter\ifx\csname l@#1\endcsname\relax
\typeout{** WARNING: IEEEtran.bst: No hyphenation pattern has been}%
\typeout{** loaded for the language `#1'. Using the pattern for}%
\typeout{** the default language instead.}%
\else
\language=\csname l@#1\endcsname
\fi
#2}}
\providecommand{\BIBdecl}{\relax}
\BIBdecl

\bibitem{robo-manip}
S.~Gu, E.~Holly, T.~Lillicrap, and S.~Levine, ``Deep reinforcement learning for
  robotic manipulation with asynchronous off-policy updates,'' in \emph{2017
  IEEE international conference on robotics and automation (ICRA)}.\hskip 1em
  plus 0.5em minus 0.4em\relax IEEE, 2017, pp. 3389--3396.

\bibitem{raza_srl}
S.~J.~A. Raza, A.~Dastider, and M.~Lin, ``Survivable hyper-redundant robotic
  arm with bayesian policy morphing,'' in \emph{2020 IEEE 16th International
  Conference on Automation Science and Engineering (CASE)}, 2020, pp. 1--7.

\bibitem{Lillicrap2016}
T.~P. Lillicrap, J.~J. Hunt, A.~Pritzel, N.~Heess, T.~Erez, Y.~Tassa,
  D.~Silver, and D.~Wierstra, ``{Continuous control with deep reinforcement
  learning},'' \emph{4th International Conference on Learning Representations,
  ICLR 2016 - Conference Track Proceedings}, 2016.

\bibitem{ac-teach}
A.~Kurenkov, A.~Mandlekar, R.~Martin-Martin, S.~Savarese, and A.~Garg,
  ``Ac-teach: A bayesian actor-critic method for policy learning with an
  ensemble of suboptimal teachers,'' \emph{arXiv preprint arXiv:1909.04121},
  2019.

\bibitem{MC-unsound}
A.~O'Hagan, ``Monte carlo is fundamentally unsound,'' \emph{The Statistician},
  pp. 247--249, 1987.

\bibitem{goal_based_planning}
D.~Verma and R.~P. Rao, ``Goal-based imitation as probabilistic inference over
  graphical models,'' in \emph{Advances in neural information processing
  systems}, 2006, pp. 1393--1400.

\bibitem{Ghavamzadeh2016}
M.~Ghavamzadeh, Y.~Engel, and M.~Valko, ``{Bayesian policy gradient},'' in
  \emph{NIPS}, vol.~17, 2016.

\bibitem{MCPG}
S.~Gu, T.~Lillicrap, Z.~Ghahramani, R.~E. Turner, and S.~Levine, ``Q-prop:
  Sample-efficient policy gradient with an off-policy critic,'' \emph{arXiv
  preprint arXiv:1611.02247}, 2016.

\bibitem{Ghavamzadeh2007}
M.~Ghavamzadeh and Y.~Engel, ``{Bayesian actor-critic algorithms},'' \emph{ACM
  International Conference Proceeding Series}, vol. 227, pp. 297--304, 2007.

\bibitem{Ghavamzadeh2015}
M.~Ghavamzadeh, S.~Mannor, J.~Pineau, and A.~Tamar, \emph{{Bayesian
  reinforcement learning: A survey}}, 2015, vol.~8, no. 5-6.

\bibitem{rtfault_diganosis}
V.~{Verma}, G.~{Gordon}, R.~{Simmons}, and S.~{Thrun}, ``Real-time fault
  diagnosis [robot fault diagnosis],'' \emph{IEEE Robotics Automation
  Magazine}, vol.~11, no.~2, pp. 56--66, 2004.

\bibitem{Bongard2006}
J.~Bongard, V.~Zykov, and H.~Lipson, ``{Resilient machines through continuous
  self-modeling},'' \emph{Science}, vol. 314, no. 5802, pp. 1118--1121, 2006.

\bibitem{Cully2015}
A.~Cully, J.~Clune, D.~Tarapore, and J.~B. Mouret, ``{Robots that can adapt
  like animals},'' \emph{Nature}, vol. 521, no. 7553, pp. 503--507, 2015.

\bibitem{Barto2019}
R.~S. Barto and A.~G., \emph{{Reinforcement learning: an introduction}}, 2019,
  vol.~53, no.~9.

\bibitem{rl_book}
R.~S. Sutton and A.~G. Barto, \emph{Reinforcement learning: An
  introduction}.\hskip 1em plus 0.5em minus 0.4em\relax MIT press, 2018.

\bibitem{thompson_sampling}
D.~J. Russo, B.~Van~Roy, A.~Kazerouni, I.~Osband, Z.~Wen \emph{et~al.}, ``A
  tutorial on thompson sampling,'' \emph{Foundations and
  Trends{\textregistered} in Machine Learning}, vol.~11, no.~1, pp. 1--96,
  2018.

\bibitem{Bayesian_dropout}
\BIBentryALTinterwordspacing
Y.~Gal and Z.~Ghahramani, ``Dropout as a bayesian approximation: Representing
  model uncertainty in deep learning,'' in \emph{Proceedings of The 33rd
  International Conference on Machine Learning}, ser. Proceedings of Machine
  Learning Research, M.~F. Balcan and K.~Q. Weinberger, Eds., vol.~48.\hskip
  1em plus 0.5em minus 0.4em\relax New York, New York, USA: PMLR, 20--22 Jun
  2016, pp. 1050--1059. [Online]. Available:
  \url{http://proceedings.mlr.press/v48/gal16.html}
\BIBentrySTDinterwordspacing

\end{thebibliography}

\end{document}